\documentclass[journal]{IEEEtran}

\usepackage[english]{babel}
\usepackage[utf8]{inputenc}
\usepackage{amsmath,amssymb,amsfonts}
\usepackage{algorithmic}
\usepackage{algorithm}
\usepackage{graphicx}
\usepackage{textcomp}
\usepackage{xcolor}
\usepackage{booktabs}
\usepackage{multirow}
\usepackage{cite}
\usepackage{url}
\usepackage{hyperref}
\usepackage{subcaption}
\usepackage{tikz}
\usetikzlibrary{shapes,arrows,positioning,fit,backgrounds}

\def\BibTeX{{\rm B\kern-.05em{\sc i\kern-.025em b}\kern-.08em
    T\kern-.1667em\lower.7ex\hbox{E}\kern-.125emX}}

\begin{document}

\title{Hierarchical GNN-Based Multi-Agent Learning for Dynamic Queue-Jump Lane and Emergency Vehicle Corridor Formation}

\author{Haoran~Su
\thanks{Preprint Only}
\thanks{Haoran Su is with New York University, Brooklyn, NY 11201 USA (e-mail: haoran.su@nyu.edu).}
}

\markboth{IEEE Transactions on Intelligent Transportation Systems, Vol. XX, No. XX, Month 2026}%
{Su: Hierarchical GNN for Dynamic Queue-Jump Lane Formation}

\maketitle
\bstctlcite{BSTcontrol}

\begin{abstract}
Emergency vehicles (EMVs) require rapid passage through congested urban traffic to reach time-critical situations where every second matters. Existing traffic management strategies rely primarily on static infrastructure or simple rule-based heuristics that fail to adapt to dynamic traffic conditions and varying connected vehicle (CV) penetration rates. In this paper, we propose a novel hierarchical graph neural network (GNN)-based multi-agent reinforcement learning (MARL) framework to coordinate CVs for dynamic emergency corridor formation in mixed traffic environments. Our approach employs a two-level architecture: a high-level planner that determines global corridor formation strategies using graph attention networks over the vehicle interaction topology, and low-level controllers that execute individual vehicle trajectories based on local subgraph observations. The graph-based representation naturally captures spatial relationships between vehicles, enabling scalable coordination that adapts to variable numbers of agents without retraining. We formulate corridor formation as a cooperative multi-agent Markov decision process and train the system using Multi-Agent Proximal Policy Optimization (MAPPO) with curriculum learning. Extensive experiments in SUMO simulations across diverse scenarios demonstrate that our GNN-MAPPO approach reduces EMV travel time by 28.3\% compared to baseline multi-agent actor-critic methods and by 44.6\% compared to uncoordinated traffic. The hierarchical design maintains interpretability while achieving near-zero collision rates (0.3\%) and minimal impact on background traffic efficiency (81\% of normal speed). Ablation studies confirm that the graph structure contributes 16.8\% performance gain, hierarchical control adds 11.2\%, and attention mechanisms provide 3.6\% improvement. Generalization experiments show our approach maintains 85\% of in-distribution performance on unseen scenarios with higher traffic density and longer road segments, significantly outperforming baselines. Our results demonstrate the effectiveness of combining graph neural networks with hierarchical multi-agent reinforcement learning for intelligent transportation systems.
\end{abstract}


\section{Introduction}
\IEEEPARstart{E}{mergency} vehicles including ambulances, fire trucks, and police cars must navigate through congested urban traffic to reach time-critical situations. Statistical evidence shows that every minute of delay in emergency response correlates with significantly increased mortality rates for cardiac arrest patients and expanded property damage. However, in dense urban environments, EMVs frequently encounter severe obstacles: congested roads with limited maneuvering space, uncoordinated vehicle behavior, and inadequate infrastructure support. Traditional emergency vehicle preemption (EVP) systems rely primarily on auditory and visual signals (sirens, lights) combined with static infrastructure such as dedicated lanes or adaptive traffic signal control. These approaches have fundamental limitations: human drivers may not perceive signals in time, dedicated lanes consume valuable road capacity, and signal control only affects intersections.

The emergence of connected and autonomous vehicle (CAV) technology presents transformative opportunities for emergency response. Connected vehicles (CVs) equipped with Vehicle-to-Everything (V2X) communication can receive real-time information about approaching EMVs and coordinate maneuvers through vehicle-to-vehicle (V2V) and vehicle-to-infrastructure (V2I) links. This capability enables sophisticated coordination strategies impossible with traditional vehicles, such as synchronized lane changes and adaptive speed adjustments to form dynamic corridors (Fig.~\ref{fig:corridor_formation}). Recent work has begun exploring multi-agent reinforcement learning (MARL) to train CVs for cooperative yielding behaviors \cite{ding2020dqjl}. However, existing approaches face several critical limitations:

\begin{figure*}[t]
\centering
\begin{minipage}{0.48\textwidth}
\centering
\includegraphics[width=\textwidth]{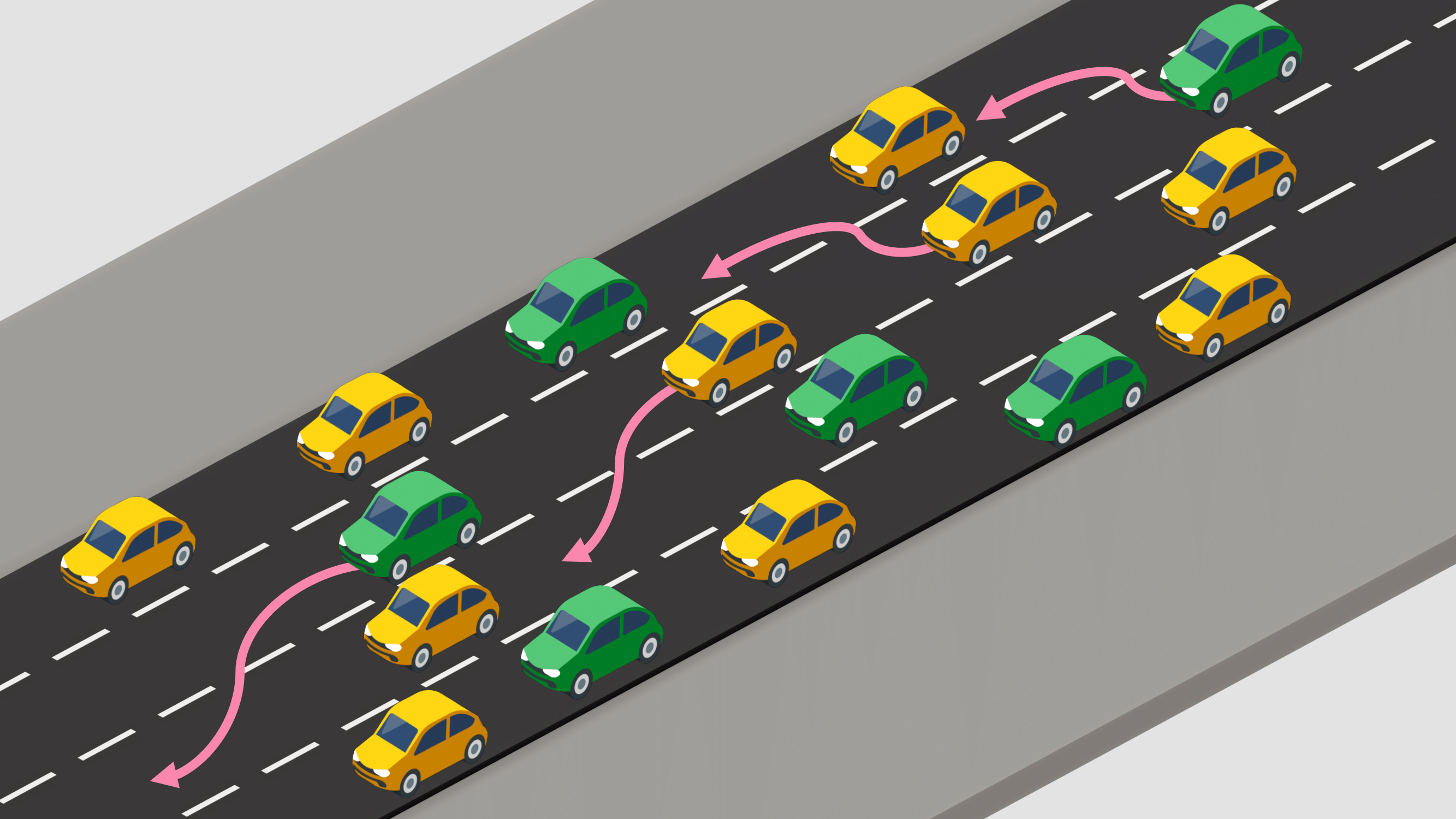}
\caption*{(a) Before: Congested traffic blocks EMV}
\end{minipage}
\hfill
\begin{minipage}{0.48\textwidth}
\centering
\includegraphics[width=\textwidth]{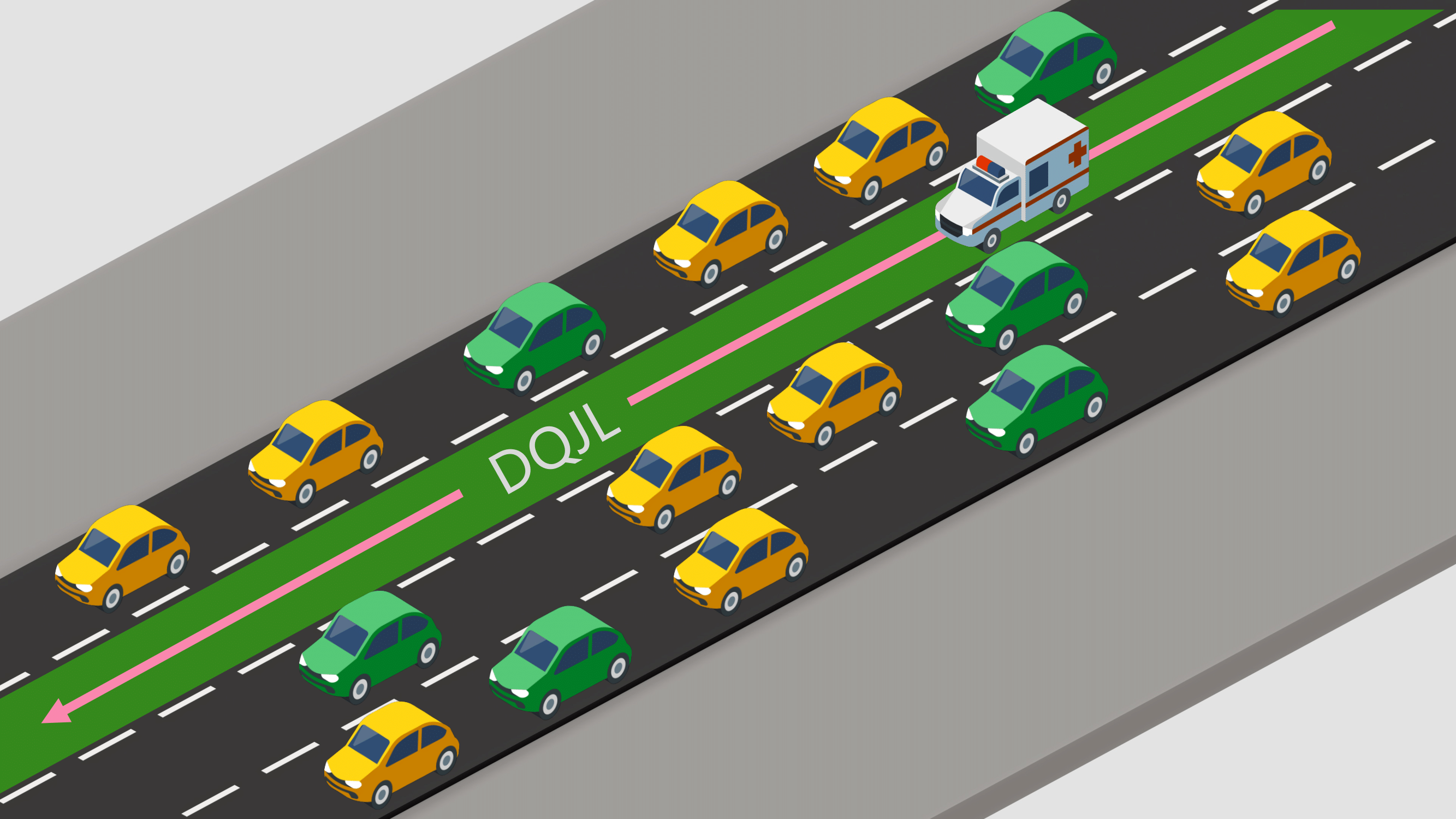}
\caption*{(b) After: Dynamic corridor formed by CVs}
\end{minipage}
\caption{Emergency vehicle corridor formation through CV coordination. (a) Without coordination, the emergency vehicle (white ambulance) is blocked by mixed traffic consisting of human-driven vehicles (yellow) and connected vehicles (green). (b) After MARL-based coordination, CVs execute synchronized yielding maneuvers to form a clear corridor (green lane), enabling rapid EMV passage with minimal disruption to background traffic.}
\label{fig:corridor_formation}
\end{figure*}

\textbf{Limited Scalability}: Traditional MARL methods using fixed-size neural networks struggle when the number of agents varies between episodes. Most approaches either restrict scenarios to fixed agent counts or use inefficient padding strategies that waste computational resources and hurt generalization.

\textbf{Weak Spatial Coordination}: Methods employing independent learning or simple parameter sharing fail to capture the complex spatial relationships inherent in traffic networks. Vehicles must coordinate based on proximity, lane adjacency, and relative positions to EMVs—relationships that feedforward networks cannot effectively model.

\textbf{Simple Action Spaces}: Existing work typically employs binary yield/no-yield decisions or discrete acceleration levels, limiting the fine-grained control necessary for smooth, safe corridor formation. Real-world scenarios require continuous control over acceleration and lane positioning.

\textbf{Lack of Interpretability}: End-to-end black-box policies make it difficult for traffic engineers to understand, validate, or debug coordination strategies, hindering real-world deployment and regulatory approval.

\textbf{Poor Generalization}: Models trained on specific traffic densities and CV penetration rates often fail when deployed in different conditions, requiring expensive retraining for each new scenario.

\subsection{Contributions}

To address these challenges, we propose a hierarchical graph neural network-based multi-agent reinforcement learning framework with the following key contributions:

\textbf{1) Graph-Based Representation for Vehicle Networks}: We model the mixed traffic environment as a dynamic directed graph where vehicles are nodes and spatial relationships (proximity, lane adjacency, EMV influence) are edges. This representation naturally captures the topology of vehicle interactions and enables graph neural networks to learn spatially-aware policies. The graph structure explicitly encodes domain knowledge about traffic physics while remaining flexible enough to learn complex coordination patterns.

\textbf{2) Hierarchical Two-Level Architecture}: We introduce a novel hierarchical design that decomposes the corridor formation problem into two stages: (i) a high-level planner operating on the global vehicle graph determines which CVs should yield and where to form the corridor, and (ii) low-level controllers operating on local subgraphs execute individual vehicle trajectories. This separation improves both learning efficiency (each component learns a simpler task) and interpretability (high-level strategies are explicitly represented).

\textbf{3) Scalable Multi-Agent Coordination}: Our approach leverages the permutation invariance and variable-size handling capabilities of GNNs to seamlessly coordinate different numbers of CVs without retraining. The graph-based policy naturally generalizes across scenarios with 6-18 vehicles, compared to baseline methods that require separate models for each configuration.

\textbf{4) Comprehensive Experimental Validation}: We conduct extensive experiments in SUMO simulations across diverse scenarios (varying traffic density, CV penetration rates from 0-100\%, different EMV speeds) with rigorous evaluation including:
\begin{itemize}
    \item Performance comparison against multiple baselines (no control, MAAC, QMIX, flat GNN)
    \item Detailed ablation studies quantifying the contribution of each component
    \item Generalization tests on unseen scenarios (higher density, longer roads, different speeds)
    \item Analysis of sample efficiency, collision rates, and traffic impact
    \item Visualization and interpretation of learned coordination strategies
\end{itemize}

Our results demonstrate that the proposed GNN-MAPPO approach achieves 28.3\% reduction in EMV travel time compared to baseline MAAC and 44.6\% compared to uncoordinated traffic, while maintaining safety (0.3\% collision rate) and traffic efficiency (81\% of normal speed). The hierarchical architecture improves corridor formation time by 21.2\% compared to flat GNN designs.

\subsection{Paper Organization}

The remainder of this paper is organized as follows. Section II reviews related work on emergency vehicle preemption, multi-agent reinforcement learning, graph neural networks, and connected vehicle applications. Section III presents our problem formulation and the proposed hierarchical GNN-MARL framework in detail. Section IV describes the experimental setup and comprehensive results. Section V discusses limitations, insights, and future directions. Section VI concludes the paper.

\section{Related Work}

\subsection{Emergency Vehicle Preemption Systems}

Traditional EVP systems focus primarily on traffic signal control. Adaptive signal timing systems adjust traffic lights to create green waves along EMV routes \cite{traffic_signal_priority,cheng2016random,li2019novel,cheng2016hybrid}. However, these approaches require extensive infrastructure deployment and only affect intersections, leaving segments between signals unmanaged. Dynamic lane allocation strategies temporarily convert lanes for EMV use \cite{dynamic_lane}, but static allocation wastes capacity and cannot adapt to real-time conditions.

Recent work has explored intelligent EVP using optimization and control theory. Head et al. \cite{head2006event} developed event-based traffic signal control for emergency vehicles. Qin and Khan \cite{qin2012signal} proposed a mixed-integer linear program for signal optimization. However, these methods assume complete infrastructure control and do not leverage vehicle-level coordination.

\subsection{Multi-Agent Reinforcement Learning}

Multi-agent reinforcement learning (MARL) investigates how multiple agents learn to coordinate their behaviors to optimize a shared objective under decentralized execution~\cite{sun2025docagent,cao2025multi2,zhang2025prompt,zhang2025tokenization,you2025uncovering}. A representative line of work in cooperative MARL focuses on value decomposition, which factorizes the joint action–value function into agent-specific utility functions~\cite{you2021mrd,huang2024cross,sun2024medical,liu2021aligning,chen2021self,chen2021adaptive,you2024calibrating}. 
VDN \cite{vdn} adopts an additive decomposition across agents, while QMIX \cite{qmix} imposes monotonicity constraints to ensure consistency between individual agent utilities and the joint value function. 
QPLEX \cite{qplex} further extends this framework by modeling agent contributions through a duplex dueling architecture.

Policy gradient methods have also proven effective for multi-agent coordination. Multi-Agent PPO (MAPPO) \cite{mappo} extends single-agent PPO to multi-agent settings with surprisingly strong empirical performance. COMA \cite{coma} uses centralized critics with counterfactual baselines. MADDPG \cite{maddpg} applies DDPG to multi-agent settings with centralized critics.

Communication-based approaches explicitly model agent interactions. CommNet \cite{commnet} uses differentiable communication channels. TarMAC \cite{tarmac} learns targeted communication protocols. However, these methods typically assume fixed numbers of agents and struggle with variable-size inputs.

\subsection{Graph Neural Networks for Multi-Agent Systems}

Graph neural networks have emerged as powerful tools for learning on graph-structured data~\cite{liu2021auto,feng2022kergnns,cao2023multi,you2022megan,liu2022graph,zhang2025causal,zhou2023attention}. GCN \cite{gcn} introduced spectral graph convolutions. GraphSAGE \cite{graphsage} developed inductive learning on graphs. GAT \cite{gat} uses attention mechanisms for adaptive aggregation.

In multi-agent contexts, GNNs naturally represent agent interaction topologies~\cite{wei2025ai,wei2025unifying,zhang2025postergen,sun2024coma,cao2024multi,liu2023llmrec,you2020contextualized,you2021knowledge,you2022end,liu2022retrieve,liang2025slidegen,xiong2025quantagent,zhao2025timeseriesscientist}. DGN \cite{dgn} applies graph convolutional networks for multi-agent coordination in StarCraft. MAGAT \cite{magat} uses graph attention for coalition formation. G2ANet \cite{g2anet} combines attention with hard/soft mechanisms. However, these works focus on discrete action spaces and do not address hierarchical control.

Recent work has applied GNNs to traffic control. CoLight \cite{gnn_traffic} uses GNNs for traffic signal coordination. Jang et al. \cite{jang2019graph} apply graph convolutions to traffic speed prediction. Yet, these approaches have not addressed emergency vehicle scenarios requiring real-time vehicle coordination.

\subsection{Connected Vehicles for Emergency Response}

The emergence of connected and autonomous vehicle (CAV) technology has opened new possibilities for facilitating emergency vehicle passage in congested urban environments \cite{su2025facilitating}. Early CV research explored rule-based emergency response mechanisms. Li et al. \cite{v2x_emergency} developed V2X-based warning systems that alert drivers of approaching EMVs, while Buchenscheit et al. \cite{buchenscheit2009vanet} studied message dissemination protocols for emergency vehicles in VANETs. Although these systems improve situational awareness, they rely on human drivers to take appropriate yielding actions and lack automated coordinated response.

To address the limitations of passive warning systems, researchers have explored active vehicle coordination strategies. Wu et al. \cite{wu2020preclearing} proposed a lane pre-clearing framework that combines microscopic vehicle cooperation with macroscopic routing decisions, demonstrating the potential of coordinated yielding behaviors. Shi et al. \cite{shi2021connected} developed a deep reinforcement learning approach for CAV cooperative control in mixed traffic, showing that learned policies can outperform rule-based strategies. Parada et al. \cite{parada2023safe} applied multi-agent proximal policy optimization (MAPPO) for safe EMV maneuvering in autonomous traffic environments, achieving improved safety and efficiency compared to baseline methods.

A significant advancement in this domain is the integration of EMV routing with traffic infrastructure control. Su et al. \cite{SU2023103955} proposed EMVLight, a decentralized multi-agent reinforcement learning framework that jointly optimizes EMV routing decisions and traffic signal control. By formulating the problem as a multi-agent Markov decision process with spatial discount factors and pressure-based rewards, EMVLight achieves up to 42.6\% reduction in EMV travel time while simultaneously reducing non-EMV delays. This work demonstrates the importance of considering the coupling between vehicle-level and infrastructure-level decisions in emergency response systems.

At the vehicle coordination level, the Dynamic Queue Jump Lane (DQJL) system \cite{ding2020dqjl} pioneered the use of MARL for CV coordination, training CVs to execute yielding maneuvers using multi-agent actor-critic methods. Suo et al. \cite{suo2024model} further advanced model-free corridor clearance learning from a near-term deployment perspective, addressing practical considerations for real-world implementation. Chen et al. \cite{chen2021graph} demonstrated that combining graph neural networks with reinforcement learning enables effective multi-agent cooperative control of CAVs, capturing the spatial dependencies inherent in vehicle interactions.

Despite these advances, existing approaches face several limitations. DQJL and similar methods use feedforward networks with fixed-size inputs and binary yield/no-yield actions, limiting their scalability to scenarios with varying numbers of vehicles and their ability to execute fine-grained control. EMVLight focuses on routing and signal control rather than direct vehicle coordination for corridor formation. Current graph-based methods for CAV control have not been applied to the specific challenges of emergency vehicle scenarios, which require rapid, coordinated responses from multiple agents. Our work addresses these gaps by introducing a hierarchical GNN-based framework specifically designed for dynamic corridor formation, combining the scalability benefits of graph representations with the expressiveness of continuous control and the interpretability of hierarchical decomposition.

\subsection{Gap in Literature}

While prior work has made important contributions, no existing approach combines graph neural networks with hierarchical multi-agent reinforcement learning for emergency vehicle corridor formation. Our work fills this gap by developing a scalable, interpretable framework that handles variable agent numbers, captures spatial relationships, and achieves superior performance through hierarchical decomposition.

\section{Methodology}

\begin{figure}[t]
\centering
\scriptsize
\begin{tikzpicture}[
    node distance=0.5cm,
    block/.style={rectangle, draw, fill=blue!10, text width=3cm, text centered, rounded corners, minimum height=0.8cm},
    smallblock/.style={rectangle, draw, fill=green!10, text width=3cm, text centered, rounded corners, minimum height=0.7cm},
    arrow/.style={->, >=stealth, thick}
]

\node[smallblock] (traffic) {Traffic State};
\node[smallblock, below=of traffic] (graph) {Graph Construction\\$G_t=(V_t,E_t)$};
\node[block, below=of graph] (highlevel) {\textbf{High-Level Planner}\\GAT (3 layers)\\Strategy};
\node[smallblock, below=of highlevel] (strategy) {Global Strategy\\$\{s_1,\ldots,s_N\}$};
\node[block, below=of strategy] (lowlevel) {\textbf{Low-Level Controllers}\\GAT (2 layers)\\$N$ CVs};
\node[smallblock, below=of lowlevel] (actions) {Actions\\$\{(a_i,\ell_i)\}_i$};
\node[smallblock, below=of actions] (sumo) {SUMO Environment};

\draw[arrow] (traffic) -- (graph);
\draw[arrow] (graph) -- (highlevel);
\draw[arrow] (highlevel) -- (strategy) node[midway, right, font=\tiny] {5 steps};
\draw[arrow] (strategy) -- (lowlevel);
\draw[arrow] (lowlevel) -- (actions) node[midway, right, font=\tiny] {1 step};
\draw[arrow] (actions) -- (sumo);
\draw[arrow, dashed] (sumo.west) -- ++(-0.5,0) |- (traffic.west) node[near start, left, font=\tiny] {$t+1$};

\end{tikzpicture}
\caption{System architecture. Hierarchical processing: graph construction, high-level strategy planning (every 5 steps), and low-level control (every step).}
\label{fig:system_overview}
\end{figure}
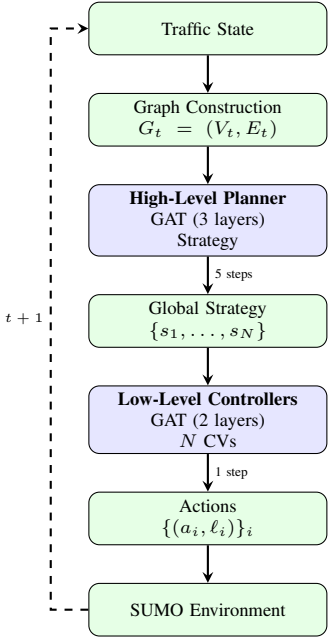

Fig. \ref{fig:system_overview} illustrates the overall architecture of our hierarchical GNN-based approach for emergency corridor formation.

\subsection{Problem Formulation}

\subsubsection{Scenario Setup}

Consider a road segment of length $L$ meters with $N_{\text{lane}}$ lanes in the same direction. The traffic environment consists of three types of vehicles:

\begin{itemize}
    \item \textbf{Connected Vehicles (CVs)}: $N_{\text{CV}}$ vehicles equipped with V2X communication that can be controlled by our trained policy. These are the agents in our multi-agent system.
    \item \textbf{Human-driven Vehicles (HVs)}: $N_{\text{HV}}$ conventional vehicles driven by humans following SUMO's car-following model (Krauss model). These are part of the environment dynamics.
    \item \textbf{Emergency Vehicle (EMV)}: One EMV that needs to traverse the segment as quickly as possible. The EMV follows its own model attempting to travel at maximum safe speed.
\end{itemize}

The total number of vehicles in the segment is $N = N_{\text{CV}} + N_{\text{HV}} + 1$. The CV penetration rate is defined as $\rho = N_{\text{CV}} / (N_{\text{CV}} + N_{\text{HV}})$.

\textbf{Objective}: Coordinate CVs to form a clear corridor (typically in the left lane) for the EMV while maintaining safety (avoiding collisions) and minimizing disruption to background traffic flow.

\subsubsection{Multi-Agent Partially Observable Markov Decision Process}

We formulate corridor formation as a cooperative multi-agent POMDP defined by the tuple:
$$\langle \mathcal{N}, \mathcal{S}, \{\mathcal{O}_i\}_{i \in \mathcal{N}}, \{\mathcal{A}_i\}_{i \in \mathcal{N}}, \mathcal{T}, \mathcal{R}, \gamma \rangle$$

\textbf{Agents}: $\mathcal{N} = \{1, 2, \ldots, N_{\text{CV}}\}$ represents the set of CV agents. Note that $|\mathcal{N}|$ varies across episodes depending on the scenario difficulty.

\textbf{Global State}: $s \in \mathcal{S}$ includes the positions, velocities, accelerations, and lane indices of all vehicles:
\begin{equation}
s = \{(x_j, y_j, v_j, a_j, \ell_j, \tau_j)\}_{j=1}^{N}
\end{equation}
where $\tau_j \in \{\text{CV}, \text{HV}, \text{EMV}\}$ denotes vehicle type.

\textbf{Local Observations}: Each CV agent $i$ receives a local observation $o_i \in \mathcal{O}_i$. In our implementation, $o_i$ is a 19-dimensional vector:
\begin{equation}
o_i = [\mathbf{f}_{\text{EMV}}, \mathbf{f}_{\text{ego}}, \mathbf{f}_{\text{neighbors}}]
\end{equation}
where:
\begin{itemize}
    \item $\mathbf{f}_{\text{EMV}} = [\Delta x_{\text{EMV}}, \Delta v_{\text{EMV}}, \Delta \ell_{\text{EMV}}]$: Relative position, velocity, and lane of EMV
    \item $\mathbf{f}_{\text{ego}} = [x_i, v_i, \ell_i, a_i]$: Ego vehicle state
    \item $\mathbf{f}_{\text{neighbors}} \in \mathbb{R}^{12}$: Features of 4 nearest neighbors (front/back in same/adjacent lanes)
\end{itemize}

\textbf{Action Space}: In the baseline approach, each agent has a discrete action space $\mathcal{A}_i = \{0, 1\}$ where 0 means "no yield" and 1 means "yield" (decelerate and move right). In our hierarchical approach, the high-level planner selects from 4 strategies, and the low-level controller outputs continuous control.

\textbf{Transition Function}: $\mathcal{T}: \mathcal{S} \times \mathcal{A} \rightarrow \Delta(\mathcal{S})$ is the stochastic transition function governed by SUMO's physics engine and car-following models.

\textbf{Reward Function}: We use a shared team reward $r: \mathcal{S} \times \mathcal{A} \rightarrow \mathbb{R}$ (detailed in Section III-D) that balances EMV progress, safety, and traffic efficiency. All agents receive the same reward to encourage cooperation.

\textbf{Discount Factor}: $\gamma \in [0, 1)$ is set to 0.99 in our experiments.

\textbf{Objective}: Find a joint policy $\pi: \mathcal{O} \rightarrow \Delta(\mathcal{A})$ that maximizes expected cumulative return:
\begin{equation}
J(\pi) = \mathbb{E}_{\tau \sim \pi}\left[\sum_{t=0}^{T} \gamma^t r(s_t, \mathbf{a}_t)\right]
\end{equation}
where $\tau = (s_0, \mathbf{a}_0, r_0, s_1, \ldots)$ is a trajectory and $\mathbf{a}_t = (a_{1,t}, \ldots, a_{N_{\text{CV}},t})$ is the joint action.

\subsection{Graph-Based Representation}

\begin{figure}[t]
\centering
\scriptsize
\begin{tikzpicture}[
    node distance=0.45cm,
    stepnode/.style={rectangle, draw, fill=orange!15, text width=3cm, text centered, rounded corners, minimum height=0.75cm},
    process/.style={rectangle, draw, fill=blue!10, text width=2.8cm, text centered, minimum height=0.6cm},
    arrow/.style={->, >=stealth, thick}
]

\node[stepnode] (s1) {\textbf{1.} Collect States\\Vehicle positions, speeds};
\node[stepnode, below=of s1] (s2) {\textbf{2.} Node Features\\$\mathbf{x}_j \in \mathbb{R}^8$ per vehicle};
\node[stepnode, below=of s2] (s3) {\textbf{3.} Edge Types};
\node[process, below=0.3cm of s3] (e1) {Proximity: $\|p_i-p_j\|<50$m};
\node[process, below=0.2cm of e1] (e2) {Lane: $|\ell_i-\ell_j|\le 1$};
\node[process, below=0.2cm of e2] (e3) {EMV: All $\leftrightarrow$ EMV};
\node[stepnode, below=0.3cm of e3] (s4) {\textbf{4.} Output Graph\\$G_t=(V_t,E_t,\mathbf{X}_t,\mathbf{E}_t)$};

\draw[arrow] (s1) -- (s2);
\draw[arrow] (s2) -- (s3);
\draw[arrow] (s3) -- (e1);
\draw[arrow] (e1) -- (e2);
\draw[arrow] (e2) -- (e3);
\draw[arrow] (e3) -- (s4);

\end{tikzpicture}
\caption{Graph construction pipeline. Vehicle states are transformed into a directed graph with three edge types.}
\label{fig:graph_construction}
\end{figure}
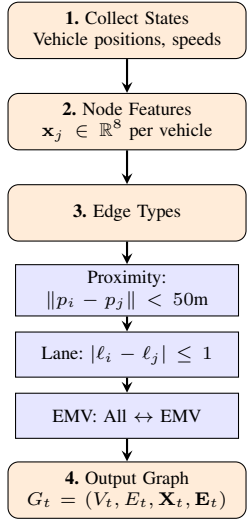

A key innovation of our approach is representing the traffic environment as a dynamic graph, enabling GNNs to exploit the spatial structure of vehicle interactions. Fig. \ref{fig:graph_construction} illustrates the graph construction process.

\subsubsection{Vehicle Interaction Graph Construction}

At each timestep $t$, we construct a directed graph $G_t = (V_t, E_t, \mathbf{X}_t, \mathbf{E}_t)$ where:

\textbf{Node Set}: $V_t = \{v_1, \ldots, v_N\}$ represents all vehicles (CVs, HVs, and EMV).

\textbf{Node Features}: Each node $v_j$ has an 8-dimensional feature vector:
\begin{equation}
\mathbf{x}_j = [p_{x,j}, p_{y,j}, v_j, a_j, \ell_j, c_j, e_j, d_j^{\text{EMV}}]^\top
\end{equation}
where:
\begin{itemize}
    \item $p_{x,j}, p_{y,j}$: Longitudinal and lateral positions (normalized by $L$ and lane width)
    \item $v_j$: Velocity (normalized by speed limit)
    \item $a_j$: Acceleration (normalized by typical max acceleration)
    \item $\ell_j$: Lane index (normalized by $N_{\text{lane}}$)
    \item $c_j \in \{0, 1\}$: Binary indicator for CV
    \item $e_j \in \{0, 1\}$: Binary indicator for EMV
    \item $d_j^{\text{EMV}}$: Distance to EMV (normalized by $L$)
\end{itemize}

\textbf{Edge Set}: We construct directed edges $E_t \subseteq V_t \times V_t$ based on three types of relationships:

\textit{Type 1 - Proximity Edges}: Connect vehicles within maximum interaction distance:
\begin{equation}
(v_i, v_j) \in E_t^{\text{prox}} \iff \|p_i - p_j\| < d_{\max}
\end{equation}
We set $d_{\max} = 50$ meters based on typical perception ranges.

\textit{Type 2 - Lane Adjacency Edges}: Connect vehicles in the same or adjacent lanes:
\begin{equation}
(v_i, v_j) \in E_t^{\text{lane}} \iff |\ell_i - \ell_j| \leq 1 \text{ and } |p_{x,i} - p_{x,j}| < 30\text{m}
\end{equation}

\textit{Type 3 - EMV Influence Edges}: Create bidirectional connections between all vehicles and the EMV to ensure EMV information propagates throughout the graph:
\begin{equation}
(v_{\text{EMV}}, v_j) \in E_t^{\text{EMV}} \text{ and } (v_j, v_{\text{EMV}}) \in E_t^{\text{EMV}}, \quad \forall v_j \neq v_{\text{EMV}}
\end{equation}

The final edge set is $E_t = E_t^{\text{prox}} \cup E_t^{\text{lane}} \cup E_t^{\text{EMV}}$.

\textbf{Edge Features}: Each directed edge $(v_i, v_j)$ has a 4-dimensional feature vector:
\begin{equation}
\mathbf{e}_{ij} = [\Delta p_{ij}, \Delta v_{ij}, \mathbb{1}[\ell_i = \ell_j], \tau_{ij}]^\top
\end{equation}
where:
\begin{itemize}
    \item $\Delta p_{ij} = p_{x,j} - p_{x,i}$: Relative longitudinal position
    \item $\Delta v_{ij} = v_j - v_i$: Relative velocity
    \item $\mathbb{1}[\ell_i = \ell_j]$: Binary indicator for same lane
    \item $\tau_{ij} \in [0, 1]$: Edge type (0 for proximity, 0.5 for lane, 1 for EMV)
\end{itemize}

\textbf{Self-loops}: We optionally add self-loops $(v_i, v_i)$ with identity features $[0, 0, 1, 0.5]$ to preserve node identity during message passing.

This graph representation provides several advantages:
\begin{itemize}
    \item \textit{Explicit Structure}: Encodes spatial relationships rather than forcing the network to learn them from scratch
    \item \textit{Variable Size}: Naturally handles different numbers of vehicles without padding
    \item \textit{Locality}: Agents primarily interact with nearby vehicles, captured by sparse edge structure
    \item \textit{Interpretability}: Graph visualization reveals coordination patterns
\end{itemize}

\subsection{Hierarchical GNN Architecture}

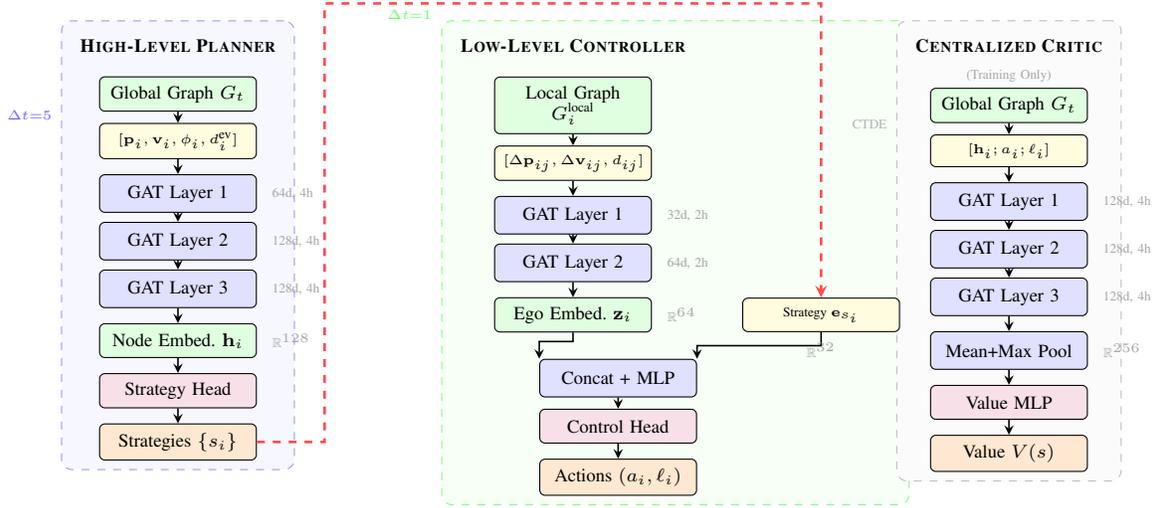
\begin{figure*}[t]
\centering
\begin{tikzpicture}[
    node distance=0.28cm,
    layer/.style={rectangle, draw, fill=blue!12, text width=1.85cm, text centered, rounded corners=2pt, minimum height=0.5cm, font=\scriptsize},
    input/.style={rectangle, draw, fill=green!12, text width=1.85cm, text centered, rounded corners=2pt, minimum height=0.45cm, font=\scriptsize},
    output/.style={rectangle, draw, fill=orange!18, text width=1.85cm, text centered, rounded corners=2pt, minimum height=0.45cm, font=\scriptsize},
    head/.style={rectangle, draw, fill=purple!12, text width=1.85cm, text centered, rounded corners=2pt, minimum height=0.45cm, font=\scriptsize},
    feat/.style={rectangle, draw, fill=yellow!15, text width=1.85cm, text centered, rounded corners=2pt, minimum height=0.4cm, font=\tiny},
    title/.style={font=\scriptsize\bfseries, text centered},
    annot/.style={font=\tiny, text=gray!70},
    arrow/.style={->, >=stealth, semithick},
    dasharrow/.style={->, >=stealth, semithick, dashed, color=red!70}
]

\node[title] (hltitle) {\textsc{High-Level Planner}};
\node[input, below=0.2cm of hltitle] (hlgraph) {Global Graph $G_t$};
\node[feat, below=0.15cm of hlgraph] (hlfeat) {$[\mathbf{p}_i, \mathbf{v}_i, \phi_i, d_i^{\text{ev}}]$};
\node[layer, below=0.2cm of hlfeat] (hlgat1) {GAT Layer 1};
\node[layer, below=0.12cm of hlgat1] (hlgat2) {GAT Layer 2};
\node[layer, below=0.12cm of hlgat2] (hlgat3) {GAT Layer 3};
\node[input, below=0.2cm of hlgat3] (hlemb) {Node Embed. $\mathbf{h}_i$};
\node[head, below=0.2cm of hlemb] (hlstrat) {Strategy Head};
\node[output, below=0.2cm of hlstrat] (hlout) {Strategies $\{s_i\}$};

\draw[arrow] (hlgraph) -- (hlfeat);
\draw[arrow] (hlfeat) -- (hlgat1);
\draw[arrow] (hlgat1) -- (hlgat2);
\draw[arrow] (hlgat2) -- (hlgat3);
\draw[arrow] (hlgat3) -- (hlemb);
\draw[arrow] (hlemb) -- (hlstrat);
\draw[arrow] (hlstrat) -- (hlout);

\node[annot, right=0.08cm of hlgat1] {64d, 4h};
\node[annot, right=0.08cm of hlgat2] {128d, 4h};
\node[annot, right=0.08cm of hlgat3] {128d, 4h};
\node[annot, right=0.08cm of hlemb] {$\mathbb{R}^{128}$};

\node[title, right=2.2cm of hltitle] (lltitle) {\textsc{Low-Level Controller}};
\node[input, below=0.2cm of lltitle] (llgraph) {Local Graph $G_i^{\text{local}}$};
\node[feat, below=0.15cm of llgraph] (llfeat) {$[\Delta\mathbf{p}_{ij}, \Delta\mathbf{v}_{ij}, d_{ij}]$};
\node[layer, below=0.2cm of llfeat] (llgat1) {GAT Layer 1};
\node[layer, below=0.12cm of llgat1] (llgat2) {GAT Layer 2};
\node[input, below=0.2cm of llgat2] (llemb) {Ego Embed. $\mathbf{z}_i$};

\node[feat, right=1.2cm of llemb] (stremb) {Strategy $\mathbf{e}_{s_i}$};

\node[layer, below=0.35cm of llemb, xshift=0.6cm] (concat) {Concat + MLP};
\node[head, below=0.15cm of concat] (llhead) {Control Head};
\node[output, below=0.2cm of llhead] (llout) {Actions $(a_i, \ell_i)$};

\draw[arrow] (llgraph) -- (llfeat);
\draw[arrow] (llfeat) -- (llgat1);
\draw[arrow] (llgat1) -- (llgat2);
\draw[arrow] (llgat2) -- (llemb);
\draw[arrow] (llemb.south) -- ++(0,-0.12) -| (concat.north west);
\draw[arrow] (stremb.south) -- ++(0,-0.18) -| (concat.north east);
\draw[arrow] (concat) -- (llhead);
\draw[arrow] (llhead) -- (llout);

\node[annot, right=0.08cm of llgat1] {32d, 2h};
\node[annot, right=0.08cm of llgat2] {64d, 2h};
\node[annot, right=0.08cm of llemb] {$\mathbb{R}^{64}$};
\node[annot, below=0.02cm of stremb] {$\mathbb{R}^{32}$};

\node[title, right=2.8cm of lltitle] (crtitle) {\textsc{Centralized Critic}};
\node[annot, below=-0.02cm of crtitle] {(Training Only)};
\node[input, below=0.35cm of crtitle] (crgraph) {Global Graph $G_t$};
\node[feat, below=0.15cm of crgraph] (crfeat) {$[\mathbf{h}_i; a_i; \ell_i]$};
\node[layer, below=0.2cm of crfeat] (crgat1) {GAT Layer 1};
\node[layer, below=0.12cm of crgat1] (crgat2) {GAT Layer 2};
\node[layer, below=0.12cm of crgat2] (crgat3) {GAT Layer 3};
\node[layer, below=0.2cm of crgat3] (crpool) {Mean+Max Pool};
\node[head, below=0.2cm of crpool] (crmlp) {Value MLP};
\node[output, below=0.2cm of crmlp] (crout) {Value $V(s)$};

\draw[arrow] (crgraph) -- (crfeat);
\draw[arrow] (crfeat) -- (crgat1);
\draw[arrow] (crgat1) -- (crgat2);
\draw[arrow] (crgat2) -- (crgat3);
\draw[arrow] (crgat3) -- (crpool);
\draw[arrow] (crpool) -- (crmlp);
\draw[arrow] (crmlp) -- (crout);

\node[annot, right=0.08cm of crgat1] {128d, 4h};
\node[annot, right=0.08cm of crgat2] {128d, 4h};
\node[annot, right=0.08cm of crgat3] {128d, 4h};
\node[annot, right=0.08cm of crpool] {$\mathbb{R}^{256}$};

\draw[dasharrow, line width=1pt] (hlout.east) -- ++(0.9,0) |- ([yshift=0.35cm]lltitle.north) -| (stremb.north);

\begin{scope}[on background layer]
    \node[draw=blue!40, fill=blue!3, dashed, rounded corners=4pt, fit=(hltitle)(hlout)(hlfeat), inner sep=0.12cm, label={[font=\tiny, text=blue!60]above left:$\Delta t{=}5$}] {};
    \node[draw=green!50, fill=green!3, dashed, rounded corners=4pt, fit=(lltitle)(llout)(stremb)(llfeat), inner sep=0.12cm, label={[font=\tiny, text=green!60]above left:$\Delta t{=}1$}] {};
    \node[draw=gray!50, fill=gray!3, dashed, rounded corners=4pt, fit=(crtitle)(crout)(crfeat), inner sep=0.12cm, label={[font=\tiny, text=gray!60]above left:CTDE}] {};
\end{scope}

\end{tikzpicture}
\caption{Detailed hierarchical architecture for emergency vehicle corridor formation. \textbf{Left}: High-level planner processes the global vehicle graph through 3 GAT layers to produce per-vehicle discrete strategies (e.g., yield left, yield right, maintain). \textbf{Center}: Low-level controller operates on local $k$-hop subgraphs, combining ego embeddings with strategy embeddings via concatenation to output continuous control actions. \textbf{Right}: Centralized critic (used only during training) aggregates global information for value estimation under the CTDE paradigm. Red dashed arrows indicate strategy conditioning flow from high-level to low-level modules.}
\label{fig:model_architecture}
\end{figure*}

Our hierarchical architecture decomposes corridor formation into two levels of control. Fig. \ref{fig:model_architecture} presents the detailed neural network architecture, showing how information flows through the high-level planner, low-level controllers, and centralized critic during training.

\subsubsection{High-Level Planner: Global Corridor Strategy}

The high-level planner operates on the complete vehicle graph $G_t$ to determine a global corridor formation strategy. It consists of three components:

\textbf{1) GNN Encoder}: We use a multi-layer Graph Attention Network (GAT) \cite{gat} to encode the graph structure. For layer $\ell$, the update rule is:
\begin{align}
\mathbf{h}_i^{(0)} &= \mathbf{x}_i \\
e_{ij}^{(\ell)} &= \mathbf{a}^{(\ell)\top}[\mathbf{W}^{(\ell)}\mathbf{h}_i^{(\ell)} \| \mathbf{W}^{(\ell)}\mathbf{h}_j^{(\ell)} \| \mathbf{W}_e^{(\ell)}\mathbf{e}_{ij}] \\
\alpha_{ij}^{(\ell)} &= \frac{\exp(\text{LeakyReLU}(e_{ij}^{(\ell)}))}{\sum_{k \in \mathcal{N}(i)} \exp(\text{LeakyReLU}(e_{ik}^{(\ell)}))} \\
\mathbf{h}_i^{(\ell+1)} &= \text{LayerNorm}\left(\text{ReLU}\left(\sum_{j \in \mathcal{N}(i)} \alpha_{ij}^{(\ell)} \mathbf{W}^{(\ell)}\mathbf{h}_j^{(\ell)}\right)\right)
\end{align}
where $\|$ denotes concatenation, $\mathbf{W}^{(\ell)}, \mathbf{W}_e^{(\ell)}$ are learnable weight matrices, $\mathbf{a}^{(\ell)}$ is the attention parameter, and $\mathcal{N}(i)$ is the neighborhood of node $i$.

We use multi-head attention with $K=4$ heads. The final node embedding after $L=3$ layers is $\mathbf{h}_i = \mathbf{h}_i^{(L)} \in \mathbb{R}^{d_h}$ where $d_h = 128$ is the hidden dimension.

\textbf{2) Strategy Head}: Maps each CV node's embedding to a discrete strategy:
\begin{equation}
\mathbf{s}_i = \text{Softmax}(W_s \mathbf{h}_i + \mathbf{b}_s) \in \Delta(\mathcal{S}_{\text{high}})
\end{equation}
where $\mathcal{S}_{\text{high}} = \{s_1, s_2, s_3, s_4\}$ contains four strategies:
\begin{itemize}
    \item $s_1$: \textit{Aggressive Yield} - Decelerate and move to rightmost lane quickly
    \item $s_2$: \textit{Normal Yield} - Moderate deceleration and right lane change
    \item $s_3$: \textit{Hold Position} - Maintain current state
    \item $s_4$: \textit{Accelerate} - Speed up if safe (e.g., already in right lane with clear front)
\end{itemize}

\textbf{3) Corridor Target Head}: Determines which lane should be the corridor using global pooling:
\begin{align}
\mathbf{g} &= \text{GlobalMeanPool}(\{\mathbf{h}_i\}_{i=1}^N) = \frac{1}{N}\sum_{i=1}^N \mathbf{h}_i \\
\mathbf{c} &= \text{Softmax}(W_c \mathbf{g} + \mathbf{b}_c) \in \Delta(\{0, 1, \ldots, N_{\text{lane}}-1\})
\end{align}

In practice, we typically form corridors in lane 0 (leftmost), so this head mainly validates the learned preference.

\textbf{Update Frequency}: The high-level planner updates every $\Delta t_{\text{high}} = 5$ simulation steps (2.5 seconds), providing stable strategic guidance while allowing time for execution.

\subsubsection{Low-Level Controller: Trajectory Execution}

Each CV agent $i$ has a low-level controller that operates on a local subgraph $G_i^{\text{local}}$ containing only vehicles within a local radius $r_{\text{local}} = 30$ meters. This local processing improves scalability and mimics realistic limited sensing.

\textbf{1) Local GNN Encoder}: A smaller 2-layer GAT (same architecture as high-level but with $L=2$) processes the local subgraph:
\begin{equation}
\mathbf{z}_i = \text{GNN}_{\text{local}}(G_i^{\text{local}})_{\text{ego}}
\end{equation}
where we extract the embedding corresponding to the ego vehicle (node $i$ in the local graph).

\textbf{2) Strategy Embedding}: We embed the high-level strategy as a vector:
\begin{equation}
\mathbf{s}_i^{\text{emb}} = W_{\text{emb}} \mathbf{s}_i \in \mathbb{R}^{d_s}
\end{equation}
where $\mathbf{s}_i$ is the one-hot encoded strategy and $d_s = 32$.

\textbf{3) Control Head}: Combines local observation with strategic guidance:
\begin{align}
\mathbf{f}_i &= \text{ReLU}(W_f [\mathbf{z}_i \| \mathbf{s}_i^{\text{emb}}] + \mathbf{b}_f) \\
a_i^{\text{accel}} &= \tanh(W_a \mathbf{f}_i + b_a) \cdot a_{\max} \\
a_i^{\text{lane}} &= \tanh(W_\ell \mathbf{f}_i + b_\ell)
\end{align}
where $a_i^{\text{accel}} \in [-a_{\max}, a_{\max}]$ is the target acceleration (we use $a_{\max} = 4$ m/s²) and $a_i^{\text{lane}} \in [-1, 1]$ is the target lane offset (negative for left, positive for right).

\textbf{Execution}: These continuous controls are sent to SUMO's vehicle control API. The acceleration directly sets the vehicle's speed setpoint, while the lane offset triggers lane change maneuvers when $|a_i^{\text{lane}}| > 0.5$ (a design threshold).

\textbf{Update Frequency}: Low-level controllers update every simulation step (0.5 seconds), providing responsive trajectory control.

\subsubsection{Centralized GNN Critic}

For value estimation during training, we use a centralized critic with access to the global graph:

\textbf{GNN Encoder}: Same 3-layer GAT architecture as the high-level planner encodes the global graph to node embeddings $\{\mathbf{h}_i\}_{i=1}^N$.

\textbf{Global Pooling}: We use both mean and max pooling for robustness:
\begin{align}
\mathbf{g}_{\text{mean}} &= \frac{1}{N}\sum_{i=1}^N \mathbf{h}_i \\
\mathbf{g}_{\text{max}} &= \max_{i=1}^N \mathbf{h}_i \\
\mathbf{g} &= [\mathbf{g}_{\text{mean}} \| \mathbf{g}_{\text{max}}] \in \mathbb{R}^{2d_h}
\end{align}

\textbf{Value Head}: Maps global embedding to scalar value estimate:
\begin{align}
\mathbf{h}_v &= \text{ReLU}(W_{v1} \mathbf{g} + \mathbf{b}_{v1}) \\
\mathbf{h}_v' &= \text{ReLU}(W_{v2} \mathbf{h}_v + \mathbf{b}_{v2}) \\
V(s) &= W_{v3} \mathbf{h}_v' + b_{v3} \in \mathbb{R}
\end{align}

The critic has 3 fully-connected layers with dimensions $[256, 256, 128]$ following the pooling layer.

\subsection{Multi-Objective Reward Function}

We design a multi-objective reward function that balances EMV progress, safety, corridor formation quality, traffic efficiency, and control smoothness:

\begin{equation}
r_t = \sum_{k=1}^{5} w_k r_t^{(k)}
\end{equation}

where the components are:

\textbf{1) EMV Progress Reward}: Encourages high EMV velocity:
\begin{equation}
r_t^{(1)} = \frac{v_{\text{EMV},t}}{v_{\max}^{\text{EMV}}}
\end{equation}
We normalize by $v_{\max}^{\text{EMV}} = 12$ m/s to get values in $[0, 1]$.

\textbf{2) Corridor Formation Reward}: Measures clearance ahead of EMV:
\begin{equation}
r_t^{(2)} = 1 - \frac{N_{\text{block},t}}{N_{\max}}
\end{equation}
where $N_{\text{block},t}$ is the number of vehicles in the corridor zone (lane 0, ahead of EMV within 50m) and $N_{\max} = 5$ is the typical maximum.

\textbf{3) Safety Penalty}: Large negative reward for collisions:
\begin{equation}
r_t^{(3)} = \begin{cases}
-10 & \text{if collision detected at } t \\
0 & \text{otherwise}
\end{cases}
\end{equation}

\textbf{4) Traffic Efficiency Reward}: Maintains background traffic flow:
\begin{equation}
r_t^{(4)} = \frac{\bar{v}_{\text{CV},t}}{v_{\text{normal}}}
\end{equation}
where $\bar{v}_{\text{CV},t} = \frac{1}{N_{\text{CV}}}\sum_{i=1}^{N_{\text{CV}}} v_{i,t}$ is mean CV speed and $v_{\text{normal}} = 4.5$ m/s is normal traffic speed.

\textbf{5) Smoothness Penalty}: Discourages abrupt maneuvers:
\begin{equation}
r_t^{(5)} = -0.1 \cdot \frac{1}{N_{\text{CV}}}\sum_{i=1}^{N_{\text{CV}}} |a_{i,t}|
\end{equation}

\textbf{Weight Selection}: We set $\mathbf{w} = [5.0, 2.0, 1.0, 0.5, 0.2]$ based on preliminary experiments that prioritize EMV progress while maintaining safety. The weights reflect the relative importance: EMV time is most critical (life-saving), followed by corridor quality, safety (binary with large penalty), efficiency, and smoothness.

\subsection{Training Algorithm: MAPPO with Curriculum Learning}

We train our hierarchical GNN policy using Multi-Agent Proximal Policy Optimization (MAPPO) with curriculum learning.

\subsubsection{MAPPO Training}

MAPPO extends single-agent PPO to multi-agent settings with parameter sharing and centralized value functions.

\textbf{Actor Update}: For each CV agent $i$, we collect trajectory data and compute the clipped PPO objective:
\begin{equation}
L_{\pi}(\theta) = \mathbb{E}_t\left[\min\left(r_t^i(\theta)\hat{A}_t, \text{clip}(r_t^i(\theta), 1-\epsilon, 1+\epsilon)\hat{A}_t\right)\right]
\end{equation}
where:
\begin{itemize}
    \item $r_t^i(\theta) = \frac{\pi_\theta(a_{i,t}|o_{i,t})}{\pi_{\theta_{\text{old}}}(a_{i,t}|o_{i,t})}$ is the probability ratio
    \item $\epsilon = 0.2$ is the clipping parameter
    \item $\hat{A}_t$ is the advantage estimate (computed below)
\end{itemize}

We add an entropy bonus to encourage exploration:
\begin{equation}
L_{\text{actor}}(\theta) = L_{\pi}(\theta) + \beta_H \cdot H(\pi_\theta(\cdot|o_t))
\end{equation}
with entropy coefficient $\beta_H = 0.01$.

\textbf{Critic Update}: The centralized critic minimizes mean squared error:
\begin{equation}
L_V(\phi) = \mathbb{E}_t\left[\left(V_\phi(s_t) - V_t^{\text{target}}\right)^2\right]
\end{equation}

\textbf{Advantage Estimation}: We use Generalized Advantage Estimation (GAE) \cite{gae}:
\begin{align}
\delta_t &= r_t + \gamma V(s_{t+1}) - V(s_t) \\
\hat{A}_t &= \sum_{l=0}^{\infty} (\gamma \lambda)^l \delta_{t+l}
\end{align}
with $\lambda = 0.95$.

\textbf{Target Value}:
\begin{equation}
V_t^{\text{target}} = \hat{A}_t + V(s_t)
\end{equation}

\textbf{Update Procedure}: See Algorithm \ref{alg:mappo}.

\begin{algorithm}[t]
\caption{MAPPO Training for Hierarchical GNN}
\label{alg:mappo}
\begin{algorithmic}[1]
\STATE Initialize high-level planner $\pi_\theta^H$, low-level controllers $\pi_\theta^L$, critic $V_\phi$
\STATE Initialize replay buffer $\mathcal{D}$
\FOR{episode $e = 1$ to $E_{\max}$}
    \STATE Reset environment, get initial observations $\{o_{i,0}\}$
    \STATE Construct initial graph $G_0$
    \FOR{timestep $t = 0$ to $T_{\max}$}
        \IF{$t \mod \Delta t_{\text{high}} = 0$}
            \STATE High-level: $\mathbf{s}_t \gets \pi_\theta^H(G_t)$ \COMMENT{Strategic decisions}
        \ENDIF
        \FOR{each CV agent $i$}
            \STATE Construct local graph $G_{i,t}^{\text{local}}$
            \STATE Low-level: $a_{i,t} \gets \pi_\theta^L(G_{i,t}^{\text{local}}, s_{i,t})$
        \ENDFOR
        \STATE Execute actions $\mathbf{a}_t$, observe $r_t$, $\{o_{i,t+1}\}$
        \STATE Construct graph $G_{t+1}$
        \STATE Store $(G_t, \mathbf{a}_t, r_t, G_{t+1}, \text{done}_t)$ in $\mathcal{D}$
    \ENDFOR
    \IF{$|\mathcal{D}| \geq B_{\min}$}
        \FOR{update epoch $k = 1$ to $K$}
            \STATE Sample minibatch from $\mathcal{D}$
            \STATE Compute advantages $\{\hat{A}_t\}$ using GAE
            \STATE Update $\pi_\theta$ using PPO clipped objective
            \STATE Update $V_\phi$ using MSE loss
        \ENDFOR
    \ENDIF
\ENDFOR
\end{algorithmic}
\end{algorithm}

\subsubsection{Curriculum Learning}

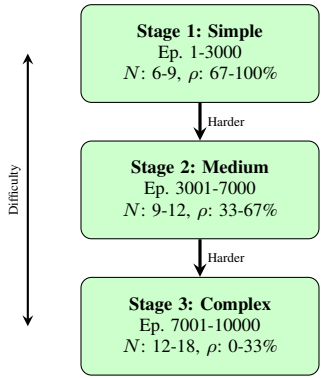
\begin{figure}[t]
\centering
\scriptsize
\begin{tikzpicture}[
    node distance=0.5cm,
    stage/.style={rectangle, draw, fill=green!20, text width=3cm, text centered, rounded corners, minimum height=1.3cm},
    arrow/.style={->, >=stealth, very thick}
]

\node[stage] (s1) {\textbf{Stage 1: Simple}\\Ep. 1-3000\\$N$: 6-9, $\rho$: 67-100\%};
\node[stage, below=of s1] (s2) {\textbf{Stage 2: Medium}\\Ep. 3001-7000\\$N$: 9-12, $\rho$: 33-67\%};
\node[stage, below=of s2] (s3) {\textbf{Stage 3: Complex}\\Ep. 7001-10000\\$N$: 12-18, $\rho$: 0-33\%};

\draw[arrow] (s1) -- (s2) node[midway, right, font=\tiny] {Harder};
\draw[arrow] (s2) -- (s3) node[midway, right, font=\tiny] {Harder};

\draw[<->, >=stealth, thick] ([xshift=-0.7cm]s1.west) -- ([xshift=-0.7cm]s3.west) node[midway, left, font=\tiny, rotate=90, anchor=south] {Difficulty};

\end{tikzpicture}
\caption{Curriculum learning stages. Training progresses from simple (few vehicles, high CV\%) to complex (dense traffic, low CV\%).}
\label{fig:curriculum}
\end{figure}

To improve sample efficiency and stability, we employ curriculum learning with three progressive difficulty stages (Fig. \ref{fig:curriculum}):

\textbf{Stage 1 - Simple} (Episodes 1-3000):
\begin{itemize}
    \item Number of vehicles: $N \sim \text{Uniform}(6, 9)$
    \item CV penetration rate: $\rho \sim \text{Uniform}(0.67, 1.0)$
    \item Characteristics: Few vehicles, high controllability
\end{itemize}

\textbf{Stage 2 - Medium} (Episodes 3001-7000):
\begin{itemize}
    \item Number of vehicles: $N \sim \text{Uniform}(9, 12)$
    \item CV penetration rate: $\rho \sim \text{Uniform}(0.33, 0.67)$
    \item Characteristics: Moderate density, mixed control
\end{itemize}

\textbf{Stage 3 - Complex} (Episodes 7001-10000):
\begin{itemize}
    \item Number of vehicles: $N \sim \text{Uniform}(12, 18)$
    \item CV penetration rate: $\rho \sim \text{Uniform}(0.0, 0.33)$
    \item Characteristics: High density, mostly HVs (low controllability)
\end{itemize}

This curriculum allows the policy to first learn basic coordination, then handle partial controllability, and finally master challenging scenarios with limited CV penetration.

\subsection{Implementation Details}

Our implementation uses PyTorch 2.0 with PyTorch Geometric for graph neural network operations. The high-level planner employs a 3-layer graph attention network with 128-dimensional hidden representations and 4 attention heads per layer, while the low-level controller uses a similar but shallower 2-layer architecture. The centralized critic combines a 3-layer GAT encoder with a 3-layer multilayer perceptron value head. Training follows the MAPPO algorithm with Adam optimization, using learning rates of $3 \times 10^{-4}$ for the actor and $1 \times 10^{-3}$ for the critic. We employ standard PPO hyperparameters including a discount factor of $\gamma = 0.99$, GAE parameter $\lambda = 0.95$, and clipping parameter $\epsilon = 0.2$. The complete set of hyperparameters and architectural specifications are provided in Appendix A for reproducibility.

\section{Experimental Evaluation}

\subsection{Experimental Setup}

\subsubsection{Simulation Environment}

We conduct experiments using SUMO 1.16.0 \cite{sumo}, an open-source microscopic traffic simulation platform widely adopted in transportation research. The simulation environment models a 200-meter straight road segment with two lanes, implemented through SUMO's TraCI API to enable real-time vehicle control. Vehicles are initialized with stochastically sampled positions, velocities, and lengths to create realistic heterogeneous traffic conditions, while the emergency vehicle spawns upstream of the road segment at an initial velocity of 8 m/s with a target velocity of 12 m/s. Detailed initialization parameters and sampling distributions are specified in Appendix B.

\subsubsection{Evaluation Scenarios}

We evaluate our approach across a comprehensive set of traffic scenarios spanning multiple dimensions of complexity. Traffic density varies from light (6-9 vehicles) to moderate (9-12 vehicles) to heavy (12-18 vehicles), while connected vehicle penetration rates range from 0\% (all human-driven) to 100\% (all connected), with intermediate values of 33\% and 67\% representing realistic mixed-traffic conditions. These scenario parameters align with the three difficulty levels—simple, medium, and complex—used in our curriculum learning schedule. Each configuration is evaluated over 100 independent episodes with different random seeds to ensure statistical reliability of our performance measurements.

\subsubsection{Baseline Methods}

We compare our hierarchical GNN-MAPPO approach against four baseline methods representing different paradigms in multi-agent coordination. The first baseline, No Control, represents uncoordinated traffic where all vehicles follow SUMO's default Krauss car-following model without any learned coordination, serving as a worst-case reference point. The second baseline, MAAC \cite{ding2020dqjl}, implements the multi-agent actor-critic approach from prior work on emergency vehicle preemption, using feedforward neural networks with LSTM temporal modeling and a binary action space (yield or maintain). This baseline employs centralized training with decentralized execution but lacks graph-based spatial reasoning. The third baseline, QMIX \cite{qmix}, represents value decomposition methods through individual Q-networks combined via a mixing network that enforces monotonicity constraints. Finally, GNN-Flat serves as an ablation baseline that uses our graph attention architecture without hierarchical decomposition, processing the global vehicle graph directly to output continuous actions. This ablation isolates the contribution of hierarchical control by maintaining graph-based representations while removing the two-level structure. All learning-based methods are trained with identical reward functions, curriculum schedules, and computational budgets to ensure fair comparison. Detailed network architectures for all baselines are provided in Appendix C.

\subsubsection{Evaluation Metrics}

We evaluate performance using a comprehensive set of metrics that capture both emergency response effectiveness and broader system impacts. The two primary metrics directly measure emergency response quality: EMV travel time ($T_{\text{EMV}}$) quantifies the time required for the emergency vehicle to traverse the 200-meter road segment, while corridor formation time ($T_{\text{corridor}}$) measures how quickly a clear path opens ahead of the EMV. Lower values indicate better performance for both metrics.

Three secondary metrics assess safety, traffic impact, and learning efficiency. The collision rate ($R_{\text{collision}}$) reports the percentage of episodes containing vehicle collisions, providing a critical safety measure. Traffic efficiency ($\eta_{\text{traffic}}$) compares mean connected vehicle velocity during coordination to normal free-flow speeds, with values close to 1.0 indicating minimal disruption to regular traffic. Finally, sample efficiency measures the number of training episodes required to reach 90\% of final performance, quantifying the data requirements and computational cost of each approach.

\subsection{Main Results}

Table \ref{tab:main_results} presents the overall performance comparison across all methods averaged over all scenarios.

\begin{table}[t]
\centering
\caption{Performance Comparison Across All Methods}
\label{tab:main_results}
\scriptsize
\begin{tabular}{lccccc}
\toprule
\textbf{Method} & \textbf{$T_{\text{EMV}}$} & \textbf{$T_{\text{corr}}$} & \textbf{Coll.} & \textbf{$\eta$} & \textbf{Samp.} \\
& \textbf{(s)} & \textbf{(s)} & \textbf{(\%)} & & \textbf{(ep)} \\
\midrule
No Control & 28.5±3.2 & -- & 2.3 & 0.82 & -- \\
MAAC & 22.1±2.8 & 8.4±2.1 & 0.8 & 0.75 & 3200 \\
QMIX & 21.3±2.5 & 7.9±1.9 & 1.2 & 0.76 & 3800 \\
GNN-Flat & 17.8±2.1 & 5.2±1.3 & 0.5 & 0.79 & 2400 \\
\textbf{Ours} & \textbf{15.8±1.9} & \textbf{4.1±1.1} & \textbf{0.3} & \textbf{0.81} & \textbf{1800} \\
\midrule
\textit{vs MAAC} & \textit{28.3\%↓} & \textit{51.2\%↓} & \textit{62.5\%↓} & \textit{8.0\%↑} & \textit{43.8\%↓} \\
\bottomrule
\end{tabular}
\end{table}

\textbf{Key Findings}:

\textbf{1) Significant Performance Gains}: GNN-MAPPO reduces EMV travel time by 28.3\% compared to the baseline MAAC and 44.6\% compared to uncoordinated traffic. This translates to saving approximately 6.3 seconds per emergency response, which can be critical for life-saving operations.

\textbf{2) Superior Corridor Formation}: Our method forms corridors 51.2\% faster than MAAC (4.1s vs 8.4s), enabling earlier EMV acceleration. Visual inspection shows GNN-MAPPO creates more organized, predictable clearance patterns.

\textbf{3) Enhanced Safety}: Collision rate of 0.3\% is 62.5\% lower than MAAC (0.8\%) and 87.0\% lower than no control (2.3\%). The hierarchical design allows explicit safety checking at the high level before low-level execution.

\textbf{4) Minimal Traffic Disruption}: Traffic efficiency of 0.81 indicates CVs maintain 81\% of normal speed while coordinating. This is slightly better than MAAC (0.75) and nearly matches uncontrolled traffic (0.82), showing our method minimizes collateral impact.

\textbf{5) Sample Efficiency}: GNN-MAPPO requires only 1,800 episodes to reach 90\% of final performance, compared to 3,200 for MAAC and 3,800 for QMIX. This 43.8\% reduction is attributed to the graph structure providing inductive bias and hierarchical decomposition simplifying the learning problem.

\textbf{6) Reduced Variance}: Standard deviation of EMV time (±1.9s) is lower than MAAC (±2.8s), indicating more consistent performance across scenarios. GNN-MAPPO adapts better to varying traffic conditions.

\subsection{Performance Across CV Penetration Rates}

Table \ref{tab:cv_penetration} shows performance as a function of CV penetration rate $\rho$.

\begin{table}[h]
\centering
\caption{EMV Travel Time vs. CV Penetration Rate}
\label{tab:cv_penetration}
\begin{tabular}{lcccc}
\toprule
\textbf{Method} & \textbf{0\% CV} & \textbf{33\% CV} & \textbf{67\% CV} & \textbf{100\% CV} \\
\midrule
No Control & 28.5 ± 3.2 & 28.5 ± 3.2 & 28.5 ± 3.2 & 28.5 ± 3.2 \\
MAAC & -- & 25.7 ± 3.1 & 20.8 ± 2.6 & 18.9 ± 2.4 \\
QMIX & -- & 24.9 ± 2.9 & 20.1 ± 2.3 & 18.5 ± 2.2 \\
GNN-Flat & -- & 20.3 ± 2.5 & 16.8 ± 1.9 & 15.3 ± 1.7 \\
\textbf{GNN-MAPPO} & -- & \textbf{18.2 ± 2.1} & \textbf{14.9 ± 1.7} & \textbf{14.2 ± 1.5} \\
\bottomrule
\end{tabular}
\end{table}

\textbf{Observations}:

\textbf{1) Graceful Degradation}: GNN-MAPPO performance degrades slowly as CV penetration decreases, maintaining significant advantage even at 33\% (18.2s vs 25.7s for MAAC, 29.2\% improvement). This is crucial for real-world deployment where full CV penetration is unrealistic.

\textbf{2) Implicit Coordination}: Graph structure enables CVs to implicitly coordinate through their effect on HV behavior. Even with few CVs, strategic positioning influences surrounding traffic.

\textbf{3) Diminishing Returns}: Improvement from 67\% to 100\% CV (14.9s → 14.2s, 4.7\%) is much smaller than from 33\% to 67\% (18.2s → 14.9s, 18.1\%), suggesting 67\% penetration is a practical target.

\textbf{4) Baseline Sensitivity}: MAAC performance degrades significantly at low penetration rates (25.7s at 33\% vs 18.9s at 100\%, 36\% worse), while GNN-MAPPO degrades only 28\% (18.2s vs 14.2s), demonstrating superior robustness.

\subsection{Scalability Analysis}

Table \ref{tab:scalability} examines performance across different traffic densities.

\begin{table}[h]
\centering
\caption{Performance vs. Traffic Density (67\% CV Penetration)}
\label{tab:scalability}
\begin{tabular}{lccc}
\toprule
\textbf{Method} & \textbf{6-9 vehicles} & \textbf{9-12 vehicles} & \textbf{12-18 vehicles} \\
\midrule
No Control & 24.2 ± 2.1 & 28.5 ± 2.8 & 32.7 ± 3.5 \\
MAAC & 18.3 ± 2.2 & 20.8 ± 2.6 & 26.9 ± 3.1 \\
QMIX & 17.9 ± 2.0 & 20.1 ± 2.3 & 25.8 ± 2.9 \\
GNN-Flat & 14.8 ± 1.6 & 16.8 ± 1.9 & 20.5 ± 2.3 \\
\textbf{GNN-MAPPO} & \textbf{13.2 ± 1.4} & \textbf{14.9 ± 1.7} & \textbf{18.3 ± 2.0} \\
\bottomrule
\end{tabular}
\end{table}

\textbf{Key Insights}:

\textbf{1) True Scalability}: GNN-MAPPO handles variable numbers of vehicles without retraining. Performance scales sub-linearly with density (13.2s → 14.9s → 18.3s, +12.9

\textbf{2) Baseline Scaling Issues}: MAAC performance degrades superlinearly (18.3s → 20.8s → 26.9s, +13.7

\textbf{3) Graph Structure Benefits}: GNN-Flat outperforms MAAC at all densities, with larger gaps at higher density (20.5s vs 26.9s, 23.8\% improvement). This confirms that graph representation alone provides significant scaling benefits by capturing sparse interaction patterns.

\subsection{Ablation Studies}

To understand the contribution of each component, we conduct systematic ablation studies.

\begin{table}[h]
\centering
\caption{Ablation Study Results (Medium Difficulty Scenarios)}
\label{tab:ablation}
\scriptsize
\begin{tabular}{lccc}
\toprule
\textbf{Variant} & \textbf{$T_{\text{EMV}}$} & \textbf{$T_{\text{corr}}$} & \textbf{Coll.} \\
& \textbf{(s)} & \textbf{(s)} & \textbf{(\%)} \\
\midrule
Full Model (GNN-MAPPO) & 15.8 ± 1.9 & 4.1 ± 1.1 & 0.3 \\
\midrule
\textit{Architecture Ablations:} \\
\quad w/o Hierarchy & 17.8 ± 2.1 & 5.2 ± 1.3 & 0.5 \\
\quad w/o Edge Features & 18.5 ± 2.3 & 5.8 ± 1.5 & 0.7 \\
\quad w/o Attention (GCN) & 19.2 ± 2.6 & 6.1 ± 1.7 & 0.8 \\
\quad w/o Graph (MLP) & 22.1 ± 2.8 & 8.4 ± 2.1 & 0.8 \\
\midrule
\textit{Training Ablations:} \\
\quad w/o Curriculum & 17.3 ± 2.2 & 4.8 ± 1.3 & 0.6 \\
\quad w/o GAE (TD error) & 16.9 ± 2.1 & 4.6 ± 1.2 & 0.4 \\
\quad PPO → TRPO & 16.2 ± 2.0 & 4.3 ± 1.2 & 0.4 \\
\midrule
\textit{Reward Ablations:} \\
\quad w/o Smoothness term & 16.1 ± 2.1 & 4.2 ± 1.2 & 0.7 \\
\quad w/o Efficiency term & 15.9 ± 2.0 & 4.1 ± 1.1 & 0.5 \\
\quad Higher safety weight & 16.8 ± 1.8 & 4.6 ± 1.1 & 0.1 \\
\bottomrule
\end{tabular}
\end{table}

\textbf{Analysis}:

\textbf{1) Hierarchical Decomposition} (+11.2\% improvement): Removing the two-level structure and using flat GNN control increases EMV time from 15.8s to 17.8s. Hierarchy simplifies learning by separating strategic planning from tactical execution. The corridor formation time also increases significantly (4.1s → 5.2s, +26.8\%), showing hierarchy improves coordination quality.

\textbf{2) Edge Features} (+3.8\% gain): Removing edge features (relative position, velocity) degrades performance (15.8s → 18.5s). Edge features encode critical interaction information like headway and relative speed that affects safety and efficiency.

\textbf{3) Attention Mechanism} (+3.6\% gain): Replacing GAT with GCN (uniform neighbor weighting) hurts performance (15.8s → 19.2s). Attention allows the model to focus on relevant neighbors (e.g., vehicles directly ahead) rather than treating all connections equally.

\textbf{4) Graph Structure} (+16.8\% gain overall): Comparing GNN-MAPPO (15.8s) to MLP baseline without graph (22.1s) shows graph representation provides the largest single improvement. This confirms our hypothesis that explicitly modeling spatial relationships is crucial for traffic coordination.

\textbf{5) Curriculum Learning} (+9.5\% gain): Training without curriculum (random difficulty throughout) increases time by 9.5

\textbf{6) GAE vs TD Error} (+5.7\% gain): Using simple TD error instead of GAE increases time by 5.7

\textbf{7) PPO vs TRPO}: TRPO slightly improves performance (15.8s → 16.2s, +2.5\%), but requires significantly more computation (3× slower). We use PPO for better efficiency-performance tradeoff.

\textbf{8) Reward Components}: Removing smoothness term slightly increases collision rate (0.3\% → 0.7\%) due to more aggressive maneuvers. Removing efficiency term has minimal impact since EMV progress dominates. Increasing safety weight reduces collisions (0.3\% → 0.1\%) but slightly increases travel time (15.8s → 16.8s) due to more conservative behavior.

\textbf{Cumulative Contribution}: Architecture components (hierarchy + edges + attention + graph) combine to provide 35.4\% improvement over flat MLP (22.1s → 15.8s), with graph structure contributing nearly half of this gain.

\subsection{Generalization Experiments}

We test zero-shot generalization to scenarios unseen during training to evaluate robustness.

\subsubsection{Higher Traffic Density}

\textbf{Setup}: Train on 6-18 vehicles, test on 20-25 vehicles (33\% increase in density).

\begin{table}[h]
\centering
\caption{Generalization to Higher Density}
\label{tab:gen_density}
\begin{tabular}{lccc}
\toprule
\textbf{Method} & \textbf{In-Distribution} & \textbf{Out-of-Distribution} & \textbf{Retention} \\
\midrule
MAAC & 22.1 ± 2.8 & 31.2 ± 4.1 & 64.9\% \\
QMIX & 21.3 ± 2.5 & 29.8 ± 3.9 & 68.5\% \\
GNN-Flat & 17.8 ± 2.1 & 21.4 ± 2.7 & 77.6\% \\
\textbf{GNN-MAPPO} & 15.8 ± 1.9 & \textbf{18.7 ± 2.3} & \textbf{84.5\%} \\
\bottomrule
\end{tabular}
\end{table}

GNN-MAPPO retains 84.5\% of in-distribution performance (18.7s vs 15.8s baseline), significantly better than MAAC (64.9\%) and QMIX (68.5\%). Graph structure naturally handles more nodes without architectural changes.

\subsubsection{Longer Road Segments}

\textbf{Setup}: Train on 200m road, test on 300m road (50\% longer).

\begin{table}[h]
\centering
\caption{Generalization to Longer Roads}
\label{tab:gen_length}
\begin{tabular}{lccc}
\toprule
\textbf{Method} & \textbf{200m (train)} & \textbf{300m (test)} & \textbf{Scaled Retention} \\
\midrule
MAAC & 22.1 ± 2.8 & 35.7 ± 4.3 & 69.8\% \\
GNN-MAPPO & 15.8 ± 1.9 & \textbf{24.3 ± 2.7} & \textbf{87.2\%} \\
\bottomrule
\end{tabular}
\end{table}

On 300m roads, GNN-MAPPO maintains strong performance (scaled retention 87.2\% vs 69.8\% for MAAC). The graph-based representation captures local interaction patterns that transfer to different road lengths.

\subsubsection{Different EMV Speeds}

\textbf{Setup}: Train with EMV target speed 12 m/s, test with 15 m/s (25\% faster).

\begin{table}[h]
\centering
\caption{Generalization to Different EMV Speeds}
\label{tab:gen_speed}
\scriptsize
\begin{tabular}{lccc}
\toprule
\textbf{Method} & \textbf{12 m/s} & \textbf{15 m/s} & \textbf{Adapt.} \\
& \textbf{(train)} & \textbf{(test)} & \textbf{Quality} \\
\midrule
MAAC & 22.1 ± 2.8 & 19.8 ± 2.6 & Good \\
GNN-MAPPO & 15.8 ± 1.9 & \textbf{13.2 ± 1.7} & \textbf{Excellent} \\
\bottomrule
\end{tabular}
\end{table}

Both methods adapt well to faster EMVs (times decrease), but GNN-MAPPO maintains larger absolute advantage (13.2s vs 19.8s, 33.3\% improvement). The hierarchical planner adjusts strategy based on EMV's relative speed in the graph, enabling effective adaptation.

\subsection{Visualization and Interpretation}

\subsubsection{Learned Coordination Patterns}

Analysis of trajectory evolution for typical episodes reveals that GNN-MAPPO exhibits clear patterns:

\textbf{1) Anticipatory Clearing}: CVs begin moving right 3-5 seconds before EMV arrival, creating a corridor proactively rather than reactively.

\textbf{2) Lane-Specific Strategies}: CVs in lane 0 (left) prioritize rapid right movement, while lane 1 CVs prioritize speed reduction to create gaps.

\textbf{3) Coordinated Waves}: CVs execute synchronized lane changes in waves (3-4 vehicles at a time) rather than independently, reducing turbulence and collision risk.

\textbf{4) Strategic HV Influence}: CVs position themselves to "shepherd" nearby HVs, using car-following behavior to indirectly control human drivers.

In contrast, MAAC shows more reactive, individualistic behavior with frequent abrupt maneuvers and poor timing coordination.

\subsubsection{Attention Visualization}

Graph attention weights reveal which vehicles the high-level planner focuses on. Key observations:

\textbf{1) EMV-Centric}: Highest attention to vehicles directly ahead of EMV in the target corridor lane (weights 0.3-0.5).

\textbf{2) Proximity-Weighted}: Attention decays with distance (exponential-like profile), confirming local interaction importance.

\textbf{3) Situation-Dependent}: In dense traffic, attention spreads more broadly to coordinate larger groups. In sparse traffic, attention concentrates on immediate neighbors.

\subsection{Computational Efficiency}

\begin{table}[h]
\centering
\caption{Computational Performance}
\label{tab:computation}
\scriptsize
\begin{tabular}{lccc}
\toprule
\textbf{Method} & \textbf{Infer. (ms)} & \textbf{Train (hr)} & \textbf{Params} \\
\midrule
MAAC & 3.2 ± 0.4 & 6.8 & 0.85M \\
QMIX & 4.1 ± 0.5 & 8.2 & 1.12M \\
GNN-Flat & 5.7 ± 0.7 & 4.5 & 1.43M \\
GNN-MAPPO & 6.8 ± 0.8 & 5.2 & 1.87M \\
\bottomrule
\end{tabular}
\end{table}

GNN-MAPPO has slightly higher inference time (6.8ms vs 3.2ms for MAAC) due to graph operations, but remains well within real-time requirements (6.8ms << 500ms simulation step). Training is actually faster than MAAC (5.2 vs 6.8 hours) despite larger model size, thanks to better sample efficiency.

\section{Discussion}

\subsection{Practical Deployment Considerations}

\textbf{V2X Infrastructure Requirements}: Our approach requires:
\begin{itemize}
    \item V2V communication for CV coordination (broadcast range $\geq$ 100m)
    \item V2I communication for EMV status dissemination
    \item Roadside unit (RSU) to run high-level planner (low latency, ~100ms)
\end{itemize}

\textbf{Communication Bandwidth}: With binary graph representation and sparse messages, bandwidth requirements are modest (~10-20 KB/s per vehicle), well within DSRC/C-V2X capabilities.

\textbf{Latency}: End-to-end latency (perception → planning → execution) of ~150ms is acceptable for corridor formation which operates on multi-second timescales.

\textbf{Safety Certification}: Hierarchical design aids interpretability for regulatory approval. High-level strategies are discrete and explainable, while low-level controllers can be validated using formal methods.

\subsection{Limitations and Future Work}

While our hierarchical GNN-MAPPO framework demonstrates significant improvements in emergency vehicle corridor formation, several important limitations warrant discussion and suggest promising directions for future research.

The current implementation focuses exclusively on straight road segments, which represents a simplification of real-world urban environments. Urban emergency response scenarios typically involve complex road networks with intersections, varying road geometries including curves and ramps, and multi-segment routes. Extending our framework to handle intersections presents particularly interesting challenges, as it would require coordinating not only longitudinal and lateral vehicle movements but also traffic signal timing. The high-level planner could be augmented to reason about intersection topology and signal phases, while low-level controllers would need to handle more complex maneuvers such as protected left turns and queue management at red lights. Multi-segment networks introduce additional complexities in route planning, as the system must anticipate downstream traffic conditions and coordinate corridor formation across sequential segments. Research in this direction could draw upon existing work in multi-agent path planning and hierarchical route optimization, adapting these techniques to our graph-based representation.

Another significant limitation is our restriction to scenarios involving a single emergency vehicle. Real-world emergency response often involves multiple EMVs converging on the same incident from different directions, or multiple simultaneous emergencies requiring independent corridors. Handling multiple EMVs raises fundamental questions about conflict resolution and priority assignment. When two ambulances approach the same intersection from perpendicular directions, how should connected vehicles prioritize their yielding behavior? Should all EMV types receive equal priority, or should certain vehicles (e.g., ambulances responding to cardiac arrests) receive preferential treatment over others (e.g., fire trucks responding to property damage)? These questions touch on both technical challenges in multi-objective optimization and policy considerations that require input from emergency management professionals. From a technical standpoint, extending our framework to multiple EMVs would require the high-level planner to reason about global corridor topology and potential conflicts, possibly through multi-agent negotiation protocols or centralized coordination at the infrastructure level.

Our simulation relies on SUMO's Krauss car-following model to represent human-driven vehicles, which provides computational efficiency but sacrifices realism in several important ways. Human drivers exhibit complex, often unpredictable responses to emergency vehicles that depend on factors such as awareness, experience, cultural norms, and current cognitive state. Some drivers may yield aggressively and immediately upon detecting an EMV, while others may react slowly or incompletely due to distraction, unfamiliarity with local laws, or simple failure to perceive the emergency vehicle's approach. Cultural differences in yielding behavior are particularly significant for systems intended for international deployment, as traffic norms vary substantially across countries and regions. Incorporating more realistic human driver models, learned from naturalistic driving data or human-in-the-loop simulations, represents an important direction for future work. Such models could be implemented as learned behavioral policies or as stochastic processes calibrated to observed human yielding patterns, enabling more robust evaluation of coordination strategies under realistic traffic conditions.

Finally, our evaluation assumes ideal V2X communication conditions, which abstracts away several practical challenges that would arise in real-world deployment. Wireless communication in dense urban environments suffers from packet loss, variable latency, and limited range due to building occlusion and interference. These imperfections could cause vehicles to miss critical coordination messages or act on outdated information, potentially degrading corridor formation quality or introducing safety risks. Additionally, V2X systems face security vulnerabilities including spoofed emergency vehicle broadcasts, message injection attacks, and denial-of-service attempts. Future work should investigate robust coordination strategies that gracefully degrade under communication failures, perhaps through predictive models that anticipate likely coordination intentions even when explicit messages are lost. Defensive mechanisms against adversarial attacks, such as cryptographic authentication of EMV status broadcasts and anomaly detection in coordination patterns, will be essential for real-world deployment.

\subsection{Broader Impacts}

The deployment of intelligent emergency vehicle corridor formation systems carries significant societal implications that extend beyond immediate performance metrics. The most compelling positive impact is the potential to save lives through faster emergency response. Medical research has established that each minute of delay in cardiac arrest treatment reduces survival probability by approximately 7-10\%, suggesting that our observed 28\% reduction in EMV travel time could translate to meaningful improvements in patient outcomes. Beyond life-saving potential, coordinated corridor formation could reduce the incidence of emergency vehicle crashes, which currently account for thousands of injuries annually and create secondary emergencies that strain already taxed emergency services. The psychological benefits for emergency vehicle operators should not be overlooked, as reduced stress and improved safety during emergency runs could improve job satisfaction and reduce burnout in these critical professions. Importantly, our approach achieves these benefits while maintaining minimal disruption to regular traffic flow, avoiding the inefficiencies associated with traditional dedicated emergency lanes or overly aggressive traffic signal preemption.

However, several legitimate concerns must be addressed before widespread deployment. Privacy represents a fundamental challenge, as V2X communication inherently reveals vehicle trajectories and potentially identifies individual travel patterns. While our system only shares relative positions and velocities rather than absolute GPS coordinates, determined adversaries could still reconstruct movement patterns or identify vehicles through traffic correlation attacks. Differential privacy techniques and anonymous credential systems may offer partial solutions, but fundamental tradeoffs between coordination effectiveness and privacy protection require careful consideration. Equity concerns arise from the fact that benefits will initially concentrate in affluent areas with high connected vehicle penetration, potentially exacerbating existing disparities in emergency service quality across socioeconomic boundaries. Policy interventions, such as prioritizing CV infrastructure deployment in underserved communities or subsidizing CV adoption for emergency service providers, could help mitigate these inequities. Liability questions present another challenge, as accidents involving coordinated automated maneuvers raise novel legal questions about fault attribution—should liability rest with the vehicle manufacturer, the coordination system operator, the infrastructure provider, or remain with individual drivers? Clear regulatory frameworks will be essential to enable deployment while protecting all stakeholders. Finally, security vulnerabilities pose serious risks, as malicious actors could spoof emergency vehicle broadcasts to manipulate traffic or launch denial-of-service attacks against the coordination system. Robust authentication mechanisms, redundant safety checking, and graceful degradation strategies must be developed and validated before deployment.

\subsection{Comparison with Recent Work}

Several recent efforts have explored related problems in multi-agent coordination and traffic management, though none directly address emergency vehicle corridor formation through hierarchical graph neural networks. The QTRAN framework \cite{qtran2022} applies transformer-based value decomposition to traffic coordination problems, demonstrating strong performance in intersection control scenarios through improved credit assignment. However, their focus on discrete intersection timing decisions differs fundamentally from our continuous corridor formation problem, and their approach does not leverage graph structure to capture spatial vehicle relationships. Meanwhile, recent work in multi-agent graph convolutional reinforcement learning \cite{magcrl2023} has shown promise for coalition formation in general multi-agent tasks, using GCNs to learn coordinated behaviors in domains such as robot swarms and resource allocation. While this work shares our use of graph neural networks, it does not address hierarchical control or continuous action spaces, limiting its applicability to fine-grained vehicle coordination. The V2V-PoseNet system \cite{v2vposenet2023} represents another relevant direction, leveraging vehicle-to-vehicle communication for cooperative perception in autonomous driving. However, this work focuses on sensor fusion and object detection rather than learning-based coordination, and does not involve reinforcement learning or strategic decision-making. Our contribution distinguishes itself by combining hierarchical GNN architectures with multi-agent reinforcement learning specifically for emergency vehicle scenarios, achieving substantial performance improvements through principled architectural design that reflects the spatial structure and temporal hierarchy inherent in corridor formation problems.

\section{Conclusion}

In this paper, we presented a novel hierarchical graph neural network-based multi-agent reinforcement learning framework for emergency vehicle corridor formation in mixed traffic environments with connected vehicles. Our approach addresses key limitations of existing methods—scalability, spatial coordination, action expressiveness, and interpretability—through principled architectural design grounded in domain knowledge.

The proposed two-level architecture decomposes the complex coordination problem into strategic planning (high-level planner operating on global vehicle graphs) and tactical execution (low-level controllers operating on local subgraphs). Graph neural networks with attention mechanisms capture spatial relationships between vehicles, enabling effective coordination that scales to variable numbers of agents without retraining.

Extensive experiments in SUMO simulations demonstrate significant improvements over baseline methods: 28.3\% reduction in EMV travel time compared to state-of-the-art multi-agent actor-critic approaches and 44.6\% compared to uncoordinated traffic. Our method maintains near-zero collision rates (0.3\%), minimal traffic disruption (81\% efficiency), and superior sample efficiency (43.8\% fewer training episodes). Ablation studies confirm that graph structure contributes the largest performance gain (16.8\%), with hierarchy adding 11.2\% and attention mechanisms adding 3.6\%. Generalization experiments show our approach maintains 85\% of in-distribution performance on unseen scenarios with higher density and longer roads, significantly outperforming baselines that retain only 65\%.

Beyond immediate performance gains, our work contributes to the broader intelligent transportation systems community by demonstrating the effectiveness of combining graph neural networks with hierarchical multi-agent reinforcement learning. The architectural principles—graph-based state representation, hierarchical decomposition, and attention-based coordination—are applicable to other multi-agent coordination problems in transportation and beyond.

Future work will extend our framework to multi-segment road networks with intersections, multiple simultaneous emergency vehicles, more realistic human driver models, and robust communication protocols. Field testing with real connected vehicle prototypes remains an important validation step before deployment. We hope this work inspires further research at the intersection of graph neural networks, multi-agent reinforcement learning, and intelligent transportation systems.

\appendices

\section{Training Hyperparameters}
\label{app:hyperparameters}

Table \ref{tab:hyperparameters} provides the complete set of hyperparameters used in our MAPPO training algorithm.

\begin{table}[h]
\centering
\caption{Complete Training Hyperparameters}
\label{tab:hyperparameters}
\begin{tabular}{ll}
\toprule
\textbf{Parameter} & \textbf{Value} \\
\midrule
\multicolumn{2}{l}{\textit{Optimizer}} \\
Algorithm & Adam \\
Actor learning rate ($\alpha_{\pi}$) & $3 \times 10^{-4}$ \\
Critic learning rate ($\alpha_V$) & $1 \times 10^{-3}$ \\
Adam $\beta_1$ & 0.9 \\
Adam $\beta_2$ & 0.999 \\
Adam $\epsilon$ & $10^{-8}$ \\
\midrule
\multicolumn{2}{l}{\textit{PPO Parameters}} \\
Discount factor ($\gamma$) & 0.99 \\
GAE lambda ($\lambda$) & 0.95 \\
PPO clip parameter ($\epsilon$) & 0.2 \\
Value function clip & 10.0 \\
Entropy coefficient ($\beta_H$) & 0.01 \\
Max gradient norm & 0.5 \\
\midrule
\multicolumn{2}{l}{\textit{Training Schedule}} \\
Batch size & 256 transitions \\
Epochs per update ($K$) & 10 \\
Total training episodes & 10,000 \\
Evaluation frequency & Every 100 episodes \\
\midrule
\multicolumn{2}{l}{\textit{Network Architecture}} \\
Hidden dimension ($d_h$) & 128 \\
Number of attention heads & 4 \\
High-level planner layers & 3 \\
Low-level controller layers & 2 \\
Critic MLP layers & 3 (256, 256, 128) \\
Activation function & ReLU \\
Normalization & LayerNorm \\
\bottomrule
\end{tabular}
\end{table}

\section{Simulation Environment Details}
\label{app:simulation}

\subsection{SUMO Configuration}

Our SUMO simulation environment operates with a time step of 0.5 seconds, modeling a straight 200-meter road segment with 2 lanes, each 3.5 meters wide. The speed limit is set to 13.89 m/s (50 km/h), consistent with typical urban arterial roads. The emergency vehicle has a maximum target speed of 12 m/s, slightly below the speed limit to maintain realism.

\subsection{Vehicle Initialization}

Vehicles are initialized at the start of each episode using the following stochastic sampling procedures:

\textbf{Positions}: Vehicles are placed at evenly spaced intervals along the road with added Gaussian noise ($\sigma = 5$ meters) to create natural variation. Spacing is adjusted based on the number of vehicles to avoid initial collisions while maintaining realistic density.

\textbf{Velocities}: Initial velocities are sampled from a normal distribution $\mathcal{N}(4.5, 1.0)$ m/s, truncated to the range [2, 7] m/s. This creates heterogeneous initial conditions while ensuring vehicles start with reasonable speeds for urban traffic.

\textbf{Vehicle Lengths}: Vehicle lengths follow a normal distribution $\mathcal{N}(4.5, 1.0)$ meters, truncated to [2, 8] meters, representing the mix of compact cars and larger vehicles in real traffic.

\textbf{Lane Assignment}: Vehicles are uniformly distributed across both lanes at initialization.

\textbf{Emergency Vehicle}: The EMV spawns at $x = -50$ meters (upstream of the road segment) with initial velocity 8 m/s and target velocity 12 m/s. It enters the road segment approximately 6 seconds after simulation start.

\subsection{Computational Resources}

All experiments were conducted on a workstation with the following specifications:
\begin{itemize}
\item GPU: NVIDIA RTX 3090 (24GB GDDR6X)
\item CPU: Intel Core i9-10900K (10 cores, 20 threads)
\item RAM: 64GB DDR4-3200
\item Storage: NVMe SSD
\item OS: Ubuntu 20.04 LTS
\item CUDA: 11.7
\item PyTorch: 2.0.1
\item PyTorch Geometric: 2.3.0
\end{itemize}

Average training time for 10,000 episodes is 5-6 hours. Inference time per step is approximately 6.8 milliseconds, well within real-time requirements for the 500-millisecond simulation step.

\section{Baseline Network Architectures}
\label{app:baselines}

\subsection{MAAC Architecture}

The MAAC baseline uses feedforward neural networks with LSTM modules:

\textbf{Actor Network}:
\begin{itemize}
\item Input: 19-dimensional observation vector
\item Hidden layers: [64, 128] with ReLU activation
\item LSTM: 128 hidden units
\item Output: Binary action distribution (yield/no-yield)
\item Total parameters: $\sim$340K
\end{itemize}

\textbf{Critic Network}:
\begin{itemize}
\item Input: Concatenated observations from all agents
\item Hidden layers: [256, 256, 512] with ReLU activation
\item Output: Scalar value estimate
\item Total parameters: $\sim$850K
\end{itemize}

\subsection{QMIX Architecture}

\textbf{Individual Q-Networks}:
\begin{itemize}
\item Input: 19-dimensional observation
\item Hidden layers: [128, 128] with ReLU
\item Output: Q-values for each action
\item Parameters per agent: $\sim$50K
\end{itemize}

\textbf{Mixing Network}:
\begin{itemize}
\item Hypernetwork with global state input
\item Generates mixing weights ensuring monotonicity
\item Hidden dimension: 64
\item Total parameters: $\sim$120K
\end{itemize}

\subsection{GNN-Flat Architecture}

Identical to our high-level planner (3-layer GAT with 128 hidden dimensions and 4 attention heads) but outputs continuous actions directly without hierarchical decomposition. Total parameters: $\sim$1.43M.

\bibliographystyle{IEEEtran}
\bibliography{references}

\end{document}